%% file: main.tex
\documentclass{article}
\usepackage[T1]{fontenc}
\usepackage{iclr2027_conference,times}
\usepackage{float}

\input{macros}

\title{Amortized Low-Rank Adaptation for Model-Based Reinforcement Learning}
\author{Fernando Palafox \& David Fridovich-Keil \\
University of Texas at Austin \\
Austin, TX, USA \\
\texttt{\{fernandopalafox,dfk\}@utexas.edu}}
\date{}

\begin{document}

\maketitle

\begin{abstract}
World models let agents plan by predicting the consequences of their actions, but changes in the environment can make them inaccurate.
We study the problem of adapting a world model to an unknown test-time environment, drawn from a known environment family, using only a few episodes of interaction.
Existing approaches trade off computational cost against expressivity, i.e., the range of models a method can produce.
For example, in-context learning is computationally cheap but limited in expressivity, and gradient-based adaptation is expressive but computationally expensive.
We present \ourmethod (\textbf{C}ontext-conditioned \textbf{L}ow-rank \textbf{A}daptation of \textbf{W}orld models), which addresses this tradeoff by using a hypernetwork to generate low-rank (LoRA) adapters at test time.
During pretraining, we simulate adaptation to a variety of environments and jointly train the hypernetwork and base world model.
At test time, we freeze the base model and use a forward pass of the hypernetwork to generate adapters from a small batch of test-time transitions.
We evaluate \ourmethod in locomotion and manipulation environment families that vary in dynamics, embodiment, and reward.
We show that, using only seconds of test-time data, \ourmethod outperforms gradient-based adaptation and in-context learning during online adaptation.
We also show that \ourmethod avoids overfitting in data-scarce regimes, that its advantage comes from the expressive adapters rather than context conditioning, and that pretraining the hypernetwork jointly with the base model outperforms training it post hoc.
\end{abstract}

\input{sections/introduction}
\input{sections/related}
\input{sections/preliminaries}
\input{sections/approach}
\input{sections/experiments}
\input{sections/conclusion}
\input{sections/acknowledgments}

\clearpage
\input{sections/ai_use}
\input{sections/ethics}
\input{sections/reproducibility}
\input{sections/appendix}

\clearpage
\bibliographystyle{iclr2027_conference}
\bibliography{references}

\end{document}

%% file: macros.tex
\usepackage{xcolor}
\usepackage{graphicx}
\usepackage{tikz}
\usetikzlibrary{arrows.meta, shapes.geometric, calc}
\usepackage{wrapfig}
\usepackage{tcolorbox}
\usepackage{xspace}
\usepackage{amsmath, mathtools, amssymb, amsthm}
\usepackage{nicefrac}
\usepackage{mathrsfs}
\usepackage{dsfont}
\usepackage[hidelinks]{hyperref}
\usepackage[capitalize,nameinlink]{cleveref}
\crefformat{equation}{(#2#1#3)}
\usepackage{booktabs}
\usepackage{tabularx}
\usepackage[margin=1cm, labelfont=bf, font=small]{caption}
\usepackage{algorithm}
\usepackage[noend]{algpseudocode}
\algrenewcommand\algorithmicrequire{\textbf{Input:}}
\algrenewcommand\algorithmicensure{\textbf{Output:}}
\usepackage{titlesec}

\titlespacing*{\paragraph}{0pt}{\parskip}{0.5em}

\definecolor{themered}{HTML}{D62728}
\colorlet{trainable}{themered!40}
\colorlet{gradflow}{themered}
\colorlet{added}{green!15}

\newcommand{\R}{\mathbb{R}}

\newcommand{\parens}[1]{\left( #1 \right)}

\newcommand{\expected}[2]{\mathbb{E}_{#1}\left[#2\right]}

\newcommand{\sg}[1]{\mathsf{sg}(#1)}

\newcommand{\env}{e}
\newcommand{\envdist}{\mathcal{E}}
\newcommand{\state}{s}

\newcommand{\wmb}{\theta_w}
\newcommand{\wma}{\phi_w}

\newcommand{\policy}{\pi}
\newcommand{\picon}{\pi_{\mathrm{con}}}
\newcommand{\piupd}{\pi_{\mathrm{upd}}}

\newcommand{\thetaWM}{\Theta_w}

\newcommand{\transition}{\zeta}
\newcommand{\traj}{\tau}
\newcommand{\trajbatch}{b}
\newcommand{\trajcon}{\trajbatch_{\mathrm{con}}}
\newcommand{\trajupd}{\trajbatch_{\mathrm{upd}}}
\newcommand{\nbatch}{N}
\newcommand{\ncon}{N_{\mathrm{con}}}
\newcommand{\nupd}{N_{\mathrm{upd}}}
\newcommand{\ntrans}{M}
\newcommand{\ntrain}{T}

\newcommand{\bufcon}{\mathcal{B}_{\mathrm{con}}}
\newcommand{\bufupd}{\mathcal{B}_{\mathrm{upd}}}

\newcommand{\z}{\mathbf{z}}
\newcommand{\rewardpred}{r}
\newcommand{\valuepred}{v}
\newcommand{\discount}{\gamma}

\newcommand{\Optim}{\mathrm{Optimizer}}

\newcommand{\Loss}{\mathcal{L}}
\newcommand{\regweight}{\lambda}

\newcommand{\new}[1]{{\setlength{\fboxsep}{1pt}\colorbox{added}{#1}}}
\newif\ifhllinenum
\algrenewcommand\alglinenumber[1]{\ifhllinenum{\setlength{\fboxsep}{1pt}\colorbox{added}{#1:}}\else#1:\fi}
\newif\ifhlfor
\algdef{SE}[FOR]{For}{EndFor}[1]{\ifhlfor\new{\textbf{for}\ #1\ \textbf{do}}\else\algorithmicfor\ #1\ \algorithmicdo\fi}{\algorithmicend\ \algorithmicfor}
\algtext*{EndFor}
\newcommand{\ourmethod}{\textsc{CLAW}\xspace}

\newcommand{\hnet}{H}
\newcommand{\hparam}{\theta_h}
\newcommand{\rank}{d_r}
\newcommand{\din}{d_{\mathrm{in}}}
\newcommand{\dout}{d_{\mathrm{out}}}
\newcommand{\hemb}{h_{\mathrm{emb}}}
\newcommand{\hcon}{h_{\mathrm{con}}}
\newcommand{\hgen}{h_{\mathrm{gen}}}

\newcommand{\cdim}{C}
\newcommand{\uemb}{u_{\mathrm{emb}}}
\newcommand{\ucon}{u_{\mathrm{con}}}
\newcommand{\chunkemb}{u_k}
\newcommand{\nchunks}{K}
\newcommand{\wmatmpl}{\phi_0}

\definecolor{draftcolor}{HTML}{1657CC}

\newtcolorbox{draftblock}{colback=draftcolor!6, colframe=draftcolor, boxrule=0.6pt, breakable, left=6pt, right=6pt, top=4pt, bottom=4pt}

%% file: sections/introduction.tex
\section{Introduction}
\label{sec:intro}

In reinforcement learning (RL), learned world models allow agents to plan by predicting the consequences of their actions \citep{ha2018world,hafner2023mastering,hansen2024tdmpc2}.
However, changes in the environment can make a world model inaccurate, and it is often desirable to adapt it without collecting large test-time datasets.
In this paper, we study the problem of adapting a world model to an unknown test-time environment drawn from a known environment family (e.g., a set of locomotion environments with varying dynamics, embodiments, or rewards).

We cast this problem as a meta-reinforcement learning problem where we amortize adaptation during pretraining, enabling online adaptation with little data.
Existing approaches face a tradeoff: they are either computationally cheap but limited in expressivity (e.g., in-context learning), or expressive but computationally expensive (e.g., gradient-based weight adaptation).
We present \ourmethod (\textbf{C}ontext-conditioned \textbf{L}ow-rank \textbf{A}daptation of \textbf{W}orld models), an adaptation method which addresses this tradeoff.
\ourmethod uses hypernetworks~\citep{ha2017hypernetworks}, neural networks trained to produce weights for other neural networks, to produce low-rank (LoRA) adapters~\citep{hu2021lora} for the world model using a small batch of test-time transitions.
This enables adaptation with the computational efficiency of in-context learning while retaining the expressivity of weight-space adaptation.

We validate \ourmethod with experiments in a variety of locomotion and manipulation environment families, where we vary dynamics, embodiments and rewards.
We show that \ourmethod outperforms gradient-based and in-context learning baselines during online adaptation, that it avoids overfitting in data-scarce regimes, that its advantage comes from the expressivity of the generated adapters (rather than context conditioning), and that pretraining the hypernetwork jointly with the base model outperforms training it post hoc.

%% file: sections/related.tex
\section{Related Work}
\label{sec:related}

\subsection{Adaptation in Model-Based RL}
\label{sec:related-adaptation}

There are three main approaches to adapt a world model to test-time environments. 
The first approach changes the model's weights directly by, e.g., gradient-based re-training \citep{levy2025meta,levy2026simulation,xu2025neural,pfrommer2020contactnets,lanier2025adapting,kim2026cosmos,wang2025latent}; the same approach, applied to policies rather than world models, also appears in model-free RL \citep{ball2023efficient,kostrikov2021offline,yin2025rapidly,amin2026recap}.
This approach is highly expressive (i.e., can produce a large range of models) at the cost of increased test-time compute and high sample complexity.

The second approach infers a learned embedding of the test-time environment (e.g., a latent task variable, a belief, or context embedding) and conditions a fixed model on it \citep{rakelly2019efficient,zintgraf2019varibad,zintgraf2018fast,duan2016rl2,mishra2017simple,james2018taskembedded,kumar2021rma,yu2017preparing,zhang2024dynamics,chen2021decision,reed2022generalist,jiang2022vima,laskin2022incontext,liu2025locoformer}.
This approach is computationally inexpensive (embedding inference is often a single forward pass of a neural network) but limited in expressivity since the weights are fixed at test time.

A third approach skips test-time adaptation entirely. 
Instead, it trains one model to be robust across an entire family of environments \citep{tobin2017domain,peng2017simtoreal,andrychowicz2020learning,lee2020learning}.
This approach does not require test-time adaptation mechanisms, but sacrifices accuracy on the test-time environment for robustness across the whole family.

Our approach, \ourmethod, is a combination of the first and second approaches. 
We use a test-time context embedding to generate low-rank adapters \citep{hu2021lora} that modify the model weights.
\ourmethod takes the best of both approaches: model adaptation is nearly instantaneous (a single forward pass of the hypernetwork), and has the expressivity of full-weight finetuning, depending on the adapter rank.

\subsection{Selecting What to Adapt}
\label{sec:related-selection}

During weight-based model adaptation, it is not always clear what weights within the model should be adapted. 
Some approaches adapt all the weights, e.g., TD-MPC2 adapts the entire model during online learning \citep{hansen2024tdmpc2}.
This has the benefit of maximum adaptation expressivity at the cost of increased test-time computation and potential model collapse \citep{smith2022legged,smith2022walk,yin2025rapidly}.
Other approaches adapt specific world model modules, e.g., the dynamics module \citep{levy2026simulation}, or specific layers, e.g., the last layers \citep{levy2025meta,harrison2018meta,davydov2024first}.
These approaches are computationally cheaper and avoid model collapse at the expense of manual module/layer selection and reduced adaptation expressivity.
A third approach does not modify weights and instead modulates the model's intermediate activations, e.g., FiLM's feature-wise affine transforms \citep{perez2017film}. 
This also avoids manual selection, but limits adaptation to modulation of existing activations rather than a full weight-space update, which is more expressive \citep{jayakumar2020multiplicative}.

Our approach is different because it inserts LoRA adapters \citep{hu2021lora} into all layers of the world model.
Therefore, we maintain most of the adaptation expressivity of full-model adaptation at a fraction of test-time compute cost.
Moreover, the pretraining procedure (\cref{alg:rollout}) implicitly teaches the hypernetwork how strongly to adapt each layer, obviating the need for manual module/layer selection.

\subsection{Hypernetworks in RL}
\label{sec:related-generating-lora}

Hypernetworks have been used in meta-RL before \citep{finn2017maml,nagabandi2018learning,raghu2019rapid} to replace test-time gradient descent with a forward pass of a neural network \citep{huang2020continual,xian2021hyperdynamics,przewilikowski2022hypermaml,beck2023hypernetworks,rezaeishoshtari2022hypernetworks,beukman2023dynamics}.
These methods often produce full weight tensors (instead of LoRA adapters).

In the domain of large language models (LLMs), existing work generates LoRA adapters from text prompts \citep{ye2021learning,mahabadi2021parameterefficient,ortizbarajas2024hyperloader,charakorn2025texttolora,chen2024generative,juki2025context,liu2026shine}.
A recent paper~\citep{bianchi2026robotic} proposes generating LoRA adapters for a robot policy from a language instruction and a demo video. 
These are all forms of amortized optimization \citep{amos2023tutorial} in which a hypernetwork is trained to amortize test-time low-rank adaptation. 

Existing hypernetwork approaches either generate full model weights (which is expensive and/or intractable for large world models), or generate adapters from explicit task descriptions for a fixed pretrained model. 
Neither approach addresses efficient world model adaptation from test-time transitions.
\ourmethod fills this gap by jointly pretraining the world model and hypernetwork to infer adapters directly from test-time interactions, enabling repeated, low-cost adaptation during deployment.

%% file: sections/preliminaries.tex
\section{Preliminaries}
\label{sec:prelim}

Each environment $\env$ is a Markov Decision Process (MDP), which we assume is drawn at test time from a uniform distribution $p(\envdist)$.
For example, $p(\envdist)$ could assign probabilities to bipedal locomotion environments that differ in actuator strength.
At each step $t$, the agent receives state $\state_t$, selects an action, and receives reward $r_t$ together with the next state $\state_{t+1}$.

We define two functions that produce the agent's actions: policies and planners.
Policies produce actions directly from states, i.e., $a_t = \policy(\state_t)$, and are often instantiated as neural networks.
In contrast, planners select an action by evaluating candidate actions using a learned model of the MDP.
That is, given world model parameters $\theta$, a planner call is given by $a_t = \policy(\state_t, \theta)$, where we overload $\policy$ for notational convenience in the next paragraph.
Planners are often instantiated as sampling-based algorithms, e.g., MPPI \citep{williams2015model}. 

We refer to the tuple $\transition_t = (\state_t, a_t, r_t, \state_{t+1})$ as a transition.
A trajectory $\traj = (\transition_0, \transition_1, \dots, \transition_{\ntrans-1})$ collects a sequence of $\ntrans$ transitions.
$p(\traj \mid \env, \policy)$ denotes the distribution over trajectories induced by acting in environment $\env$ according to the policy or planner $\policy$.
A batch of trajectories with size $\nbatch$ is the tuple $\trajbatch = (\traj_1, \dots, \traj_{\nbatch})$, where each trajectory is drawn independently and identically distributed from $p(\traj \mid \env, \policy)$.
At test time, the agent interacts with environments from $p(\envdist)$ which it may or may not have encountered during pretraining.

\subsection{World Model}
\label{sec:worldmodel}

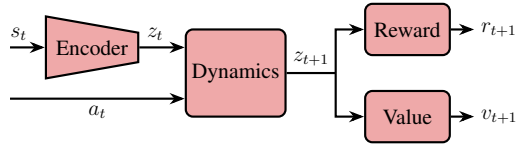
\begin{wrapfigure}[9]{r}{0.5\textwidth}
\centering
\resizebox{\linewidth}{!}{\input{figures/worldmodel}}
\vspace{-15pt}
\caption{A single forward pass of a world model used to plan by maximizing predicted return. Blocks in red have trainable parameters, which constitute $\wmb$.}
\label{fig:worldmodel}
\end{wrapfigure}

World models are learned approximations of MDPs that agents can use to plan actions. 
They often consist of an encoder, a dynamics function, a value function, and a reward function~\citep{hafner2019dream,hafner2023mastering,hansen2024tdmpc2,levy2026simulation}.
These modules are neural networks jointly trained using data from high-fidelity simulation or the real world.
We refer to the parameters for all modules as $\wmb$, the \textit{base} parameters.
In this paper, the agent uses $\wmb$ with a planner $\piupd$ to select actions, i.e., $a_t = \piupd(\state_t,\wmb)$.
\Cref{fig:worldmodel} shows the control flow for action selection given a set of candidate actions.

First, the encoder produces a latent state $z_t$ from an environment state $\state_t$.
Then, the dynamics module predicts the next latent state $z_{t+1}$ given $z_t$ and $a_t$.
Next, the reward and value modules use $z_{t+1}$ to compute the reward $\rewardpred_{t+1}$ and value $\valuepred_{t+1}$.
Finally, the planner ranks $a_t$ using the estimated return $\rewardpred_{t+1} + \discount\,\valuepred_{t+1}$, where $\discount \in [0,1)$ is the discount factor.

In this paper, we apply our method, \ourmethod, to the TD-MPC2 world model architecture~\citep{hansen2024tdmpc2} because it consistently achieves strong performance across a variety of continuous control tasks using a single set of hyperparameters.\footnote{TD-MPC2 pretrains a model-free policy to seed a sampling-based planner. To avoid confusion, we include the parameters for this policy within $\wmb$.}
We emphasize that \ourmethod is agnostic to world model architecture since our only requirement is that we can sample environments during pretraining and insert LoRA adapters into world model modules.
Therefore, \ourmethod can be applied to, e.g., action-conditioned JEPA world models~\citep{maes2026leworldmodel,assran2025v} or Dreamer-style architectures~\citep{hafner2019dream,hafner2023mastering}.

\subsection{Low-Rank Adaptation}
\label{sec:lora}

At test time, the base parameters $\wmb$ stay frozen and we insert LoRA adapters~\citep{hu2021lora} into the world model's layers.
Concretely, for an adapted layer with base weight matrix $W_0 \in \R^{\dout \times \din}$, where $\dout$ and $\din$ are the layer's output and input dimensions, the effective weight is
\begin{equation}
  W \;=\; W_0 \;+\; A B,
  \qquad
  A \in \R^{\dout \times \rank},
  \qquad
  B \in \R^{\rank \times \din},
  \qquad
  \rank \ll \min(\dout, \din).
  \label{eq:lora}
\end{equation}
The factors $A$ and $B$ constitute the adapter, and $\rank$ is the low-rank bottleneck dimension.
We define $\wma$ as the collection of all $A, B$ factors across the adapted layers.
For simplicity, we adapt all world model layers and keep rank $\rank$ constant.

\subsection{Hypernetworks}
\label{sec:hypernetworks}

A hypernetwork is a neural network $H_{\hparam}$ that outputs the weights of another network \citep{ha2017hypernetworks}.
Given an input $x$, the target network parameters $\phi$ are produced as
\begin{equation}
  \phi = H_{\hparam}(x),
  \label{eq:hypernet_general}
\end{equation}
where $\hparam$ are the weights for the hypernetwork.
In the next section, we present more details on the architecture and training procedure for our LoRA-generating hypernetwork. 

%% file: figures/worldmodel.tex
\def\figfont{\fontsize{12pt}{14pt}\selectfont}
\begin{tikzpicture}[
  block/.style={draw, rounded corners, fill=trainable, minimum width=1.65cm, minimum height=1.0cm, align=center, font=\figfont},
  dynblock/.style={draw, rounded corners, fill=trainable, minimum width=1.7cm, minimum height=1.7cm, align=center, font=\figfont},
  enc/.style={draw, trapezium, trapezium angle=100, shape border rotate=90, fill=trainable,
              minimum width=1.1cm, minimum height=1.8cm, align=center, font=\figfont},
  symlabel/.style={font=\figfont},
  ->, >=Stealth, very thick
]
  \node[enc]      (enc) at (0, 0.5)      {Encoder};
  \node[dynblock] (dyn) at (2.8, 0)      {Dynamics};
  \node[block]    (rew) at (6.15, 0.85)  {Reward};
  \node[block]    (val) at (6.15, -0.85) {Value};

  \draw (-1.6, 0.5)  -- node[above, symlabel, pos=0.25] {$\state_t$}     (enc.west);
  \draw (enc.east)   -- node[above, symlabel, pos=0.35] {$z_t$} (dyn.west |- enc.east);
  \draw (-1.6, -0.5) -- node[below, symlabel] {$a_t$}     ($(dyn.west)+(0,-0.5)$);
  \draw[-] (dyn.east) -- node[above, symlabel] {$z_{t+1}$} (4.75, 0);
  \draw (4.75, 0) |- (rew.west);
  \draw (4.75, 0) |- (val.west);
  \draw (rew.east) -- (7.45, 0.85) node[right, symlabel] {$\rewardpred_{t+1}$};
  \draw (val.east) -- (7.45, -0.85) node[right, symlabel] {$\valuepred_{t+1}$};
\end{tikzpicture}

%% file: sections/approach.tex
\section{Approach}
\label{sec:proposed}

We now present \ourmethod, a model-based RL adaptation algorithm that uses a hypernetwork to generate low-rank adapters at test time. 
This section is structured as follows: 
First, we detail the architecture of our hypernetwork (\cref{sec:hypernet}). 
Then, we describe a pretraining procedure to jointly train the base model and hypernetwork (\cref{sec:training-procedure}). 

\subsection{Hypernet-Based Adapter Generation}
\label{sec:hypernet}

\begingroup
\setlength{\textfloatsep}{8pt}
\begin{figure}[b]
\centering
\input{figures/hypernet-forward}
\vspace{-0.4cm}
\caption{Forward pass of the hypernetwork $\hnet_{\hparam}$ when generating chunk $k$ of $\Delta\wma$ using a batch $\trajcon$ of $\ncon$ trajectories, each with $\ntrans$ transitions. In the $\trajcon$ block, depth represents the batch dimension. Blocks in red have trainable parameters which constitute $\hparam$.}
\label{fig:hypernet-forward}
\end{figure}
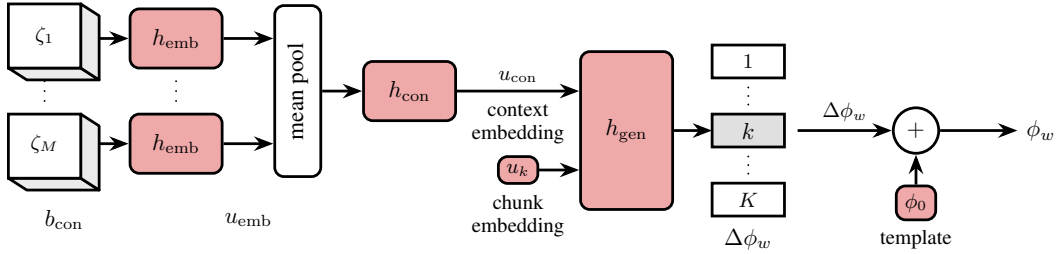
\endgroup

We use a hypernetwork to generate LoRA adapters for each environment, online at test time.
Using a context batch $\trajcon$ of $\ncon$ trajectories of $\ntrans$ transitions each, it generates LoRA adapters $\wma$ for the world model's base weights $\wmb$.
Formally:
\begin{equation}
  \wma \;=\; \hnet_{\hparam}(\trajcon).
  \label{eq:hypernet}
\end{equation}
We now describe the architecture of $\hnet$ and how $\trajcon$ is used to generate $\wma$.

First, we generate an embedding $\uemb$ for each transition $\transition \in \trajcon$ using a small multi-layer perceptron (MLP) $\hemb$.
Then, we average the embeddings and feed them into a second MLP $\hcon$ to produce a context embedding $\ucon$ that identifies the environment.
Note that embedding each transition independently, then averaging, discards information about transition order.

Once we have identified the environment with $\ucon$, we feed it into $\hgen$, an MLP, to generate $\Delta\wma$, which is added to a learned template $\wmatmpl$ to produce $\wma$, i.e.,
\begin{equation}
  \wma = \wmatmpl + \Delta\wma. 
\end{equation}
Templates have been shown to improve learning performance in hypernetworks~\citep{beck2023hypernetworks}.

Generating $\Delta\wma$ from a single forward pass would require an impractically large generator network $\hgen$.
Instead, $\hgen$ is a small network we reuse to generate $\Delta\wma$ in chunks. 
Concretely, following \citet{von2020continual}, we divide $\Delta\wma$ into $\nchunks$ chunks and generate chunk $k$ by feeding the chunk embedding $\chunkemb$ and the context embedding $\ucon$ into $\hgen$.
\Cref{fig:hypernet-forward} illustrates the full procedure.

\subsection{Amortizing Low-Rank Adaptation}
\label{sec:training-procedure}

We now describe the procedure to jointly pretrain the world model and hypernetwork, and thereby amortize test-time low-rank adaptation.
At a high level, \ourmethod wraps a standard world model pretraining algorithm with a loop that iterates between environments sampled from an environment family. 
This allows us to train the hypernetwork to generate adapters for each environment.
In the following subsections we explain: \textbf{1)} a single-environment training step for the base model and hypernetwork parameters (i.e., $\wmb$, $\hparam$), \textbf{2)} a training loss that averages over environments, thus meta-training the base and hypernetwork for online adaptation, and \textbf{3)} the full pretraining algorithm, instantiated for the TD-MPC2 architecture \citep{hansen2024tdmpc2}.

\paragraph{Single-Environment Training Step.}
\label{sec:single-step}

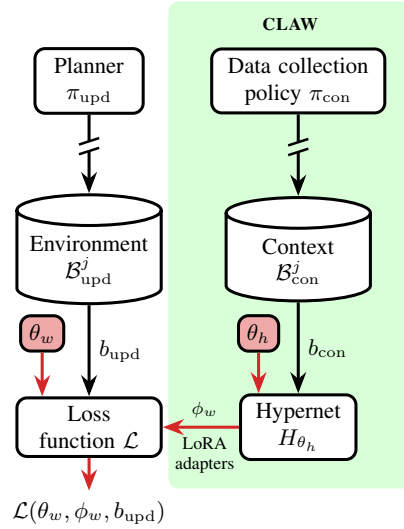
\begin{wrapfigure}{r}{0.4\textwidth}
\centering
\resizebox{\linewidth}{!}{\input{figures/forward-pass}}
\caption{
Computing the loss $\Loss$ for environment $j$ during \ourmethod pretraining. Green marks what \ourmethod adds to standard world model pretraining loss computation. Red marks trainable parameters and their gradient path, and ticks mark a stopped gradient. \ourmethod can be applied to any world model training algorithm with the same high-level computation graph.}
\label{fig:forward-pass}
\vspace{-1.25cm}
\end{wrapfigure}

Given environment $j$, we start by filling two buffers: $\bufcon^{j}$, a context buffer containing trajectories collected using a context-collection policy $\picon$, and $\bufupd^{j}$, a buffer containing trajectories collected using the agent's planner, $\piupd$.
Next, we sample a batch $\trajcon$ of $\ncon$ trajectories from $\bufcon^{j}$ and produce environment-specific low-rank adapters using the hypernetwork, i.e., $\wma = \hnet_{\hparam}(\trajcon)$.
Then, we combine $\wmb$ and $\wma$ per \cref{eq:lora},
and use a batch $\trajupd$ of $\nupd$ trajectories sampled from $\bufupd^{j}$ to compute the loss $\Loss$ (detailed in the next section).
Finally, we differentiate $\Loss$ with respect to $\wmb$ and $\hparam$, and update $\wmb$ and $\hparam$ with an Adam optimizer \citep{kingma2014adam}.

The result is a hypernetwork trained to produce low-rank adapters $\wma$ for $\wmb$ using $\trajcon$, and a base model trained to be adapted with $\wma$.
A visualization of this procedure can be seen in \cref{fig:forward-pass}.

Our method collects $\trajcon$ using the context-collection policy $\picon$ instead of the agent's planner. 
There are two reasons for this approach:
\begin{enumerate}
  \item It ensures the distribution of context trajectories is the same during training and testing, facilitating environment identification for the hypernetwork $\hnet$.
  \item In future work, we can directly optimize the policy $\picon$ to produce trajectories $\trajcon$ which are better at identifying the environment and/or lead to better test-time performance.
\end{enumerate} 
In our experiments, $\picon$ is always a random-action policy. 
This fixes context data quality across adaptation baselines so that performance differences reflect the adaptation method rather than the collected context data.
However, practitioners may choose to use a different policy, e.g., one that is safe for real-life deployment (but not necessarily high-performing).
The use of a non-random $\picon$ is fully supported by \ourmethod as long as $\picon$ can be queried during pretraining.

\paragraph{Multi-Environment Training Loss.}
\label{sec:loss}

The loss in \cref{fig:forward-pass} decomposes as 
\begin{equation}
  \Loss(\wmb, \wma, \trajupd) \;=\; \Loss_{\mathrm{TDMPC2}}(\thetaWM, \trajupd) \;+\; \regweight \|\wma\|_2^2,
  \label{eq:loss-decomp}
\end{equation}
where $\thetaWM$ are the merged base and adapter weights per \cref{eq:lora},
$\regweight \in \R_{\geq 0}$ weighs the adapter's $L_2$ penalty, and $\Loss_{\mathrm{TDMPC2}}$ is the training loss for a TDMPC2 world model described by \citet{hansen2024tdmpc2}.
If a different world model architecture is used, $\Loss_{\mathrm{TDMPC2}}$ may be replaced with the corresponding loss.
\Cref{fig:forward-pass} computes a single-environment, single-batch approximation of the objective in the following minimization problem:
\begin{equation}
  \min_{\wmb,\, \hparam} \;
  \expected{\env}{\,
    \expected{\trajcon,\, \trajupd}{\Loss}
  },
  \label{eq:objective}
\end{equation}
where $\env \sim p(\envdist)$, and $\trajcon$ and $\trajupd$ are batches of $\ncon$ and $\nupd$ trajectories, respectively. 
Each trajectory is drawn i.i.d.\ from $p(\traj \mid \env, \picon)$ and $p(\traj \mid \env, \piupd)$, respectively.

\input{algorithms/pretraining}

Problem \Cref{eq:objective} illustrates how our method can be applied to any world model architecture. 
If we remove $\trajcon$ and the $\regweight \|\wma\|_2^2$ term, the inner expectation of \cref{eq:objective} is the objective of a standard world model training problem.
Our approach, \ourmethod, defines the hypernetwork machinery (green box in \cref{fig:forward-pass}), and uses it to amortize low-rank adaptation by wrapping the standard objective with an expectation over the environment family $p(\envdist)$.

\paragraph{Applying \ourmethod to the TD-MPC2 Architecture.}
\label{sec:alg}

\Cref{alg:rollout} shows the TD-MPC2 \citep{hansen2024tdmpc2} pretraining algorithm modified to use \ourmethod.
It includes the following additions:
\textbf{1)} logic to sample a rollout environment from $p(\envdist)$ every episode, thereby approximating the outer expectation in \cref{eq:objective};
\textbf{2)} logic to sample a \textit{training} environment from $p(\envdist)$ every step, thereby stabilizing multi-environment training;
\textbf{3)} a buffer $\bufcon^j$ per environment $j$ with corresponding samples $\trajcon$,
and \textbf{4)} a forward pass of $\hnet_{\hparam}$ to produce adapters $\wma$ given $\trajcon$.

We denote additions to the original TD-MPC2 algorithm with \new{green highlights} in \cref{alg:rollout}.
Our algorithm uses a modified version of the TD-MPC2 weight update, \textsc{TDMPC2Update}, which we include as \cref{alg:tdmpc2update} in \cref{app:tdmpc2update}.
The result of \cref{alg:rollout} is a base TD-MPC2 model and hypernet that can generate test-time LoRA adapters for any layer of the base model.

We emphasize that \ourmethod can be applied regardless of the chosen world model architecture or training algorithm.
Specifically, an adapter-generating hypernetwork can be trained as long as one can sample environments, maintain a buffer for each of them, and maintain an unbroken gradient chain between the loss and hypernetwork (as shown by the red lines in \cref{fig:forward-pass}).

%% file: figures/hypernet-forward.tex
\resizebox{\linewidth}{!}{%
\begin{tikzpicture}[
  block/.style={draw, rounded corners, fill=trainable, minimum width=1.4cm, minimum height=0.9cm, align=center},
  genblock/.style={draw, rounded corners, fill=trainable, minimum width=1.4cm, minimum height=2.4cm, align=center},
  poolblock/.style={draw, rounded corners, minimum width=0.7cm, minimum height=2.6cm, align=center},
  ekblock/.style={draw, rounded corners, fill=trainable, minimum width=0.6cm, minimum height=0.4cm, align=center},
  lbl/.style={align=center},
  ->, >=Stealth, very thick
]
  \node[block] (embTop) at (3.05, 1.4)  {$\hemb$};
  \foreach \dy in {-0.15,0,0.15} {\fill (3.05, 0.6+\dy) circle (0.4pt);}
  \node[block] (embBot) at (3.05, -0.2) {$\hemb$};

  \node[poolblock] (pool) at (4.9, 0.6)  {\rotatebox{90}{mean pool}};
  \node[block] (agg)  at (6.6, 0.6)  {$\hcon$};

  \def\dx{0.28}
  \def\dy{0.18}

  \draw[fill=black!8, line join=round] (0.5,0.95) -- (1.6,0.95) -- (1.6+\dx,0.95-\dy) -- (0.5+\dx,0.95-\dy) -- cycle;
  \draw[fill=black!8, line join=round] (1.6,0.95) -- (1.6,1.925) -- (1.6+\dx,1.925-\dy) -- (1.6+\dx,0.95-\dy) -- cycle;
  \draw[-, fill=white] (0.5,0.95) rectangle (1.6,1.925);
  \node[lbl, font=\footnotesize] at (1.05, 1.44) {$\transition_1$};

  \foreach \dyy in {-0.15,0,0.15} {\fill (1.05, 0.6+\dyy) circle (0.4pt);}

  \draw[fill=black!8, line join=round] (0.5,-0.725) -- (1.6,-0.725) -- (1.6+\dx,-0.725-\dy) -- (0.5+\dx,-0.725-\dy) -- cycle;
  \draw[fill=black!8, line join=round] (1.6,-0.725) -- (1.6,0.25) -- (1.6+\dx,0.25-\dy) -- (1.6+\dx,-0.725-\dy) -- cycle;
  \draw[-, fill=white] (0.5,-0.725) rectangle (1.6,0.25);
  \node[lbl, font=\footnotesize] at (1.05, -0.2) {$\transition_{\ntrans}$};

  \node[lbl] at (1.35, -1.35) {$\trajcon$};

  \draw (1.6+\dx, 1.4)   -- (embTop.west);
  \draw (1.6+\dx, -0.2)  -- (embBot.west);

  \draw (embTop.east) -- (pool.west |- embTop);
  \draw (embBot.east) -- (pool.west |- embBot);
  \node[lbl] at (4.15, -1.35) {$\uemb$};
  \draw (pool.east) -- (agg.west);

  \node[genblock] (gen) at (9.9, 0)    {$\hgen$};

  \draw (agg.east) -- node[above, font=\footnotesize] {$\ucon$} ($(gen.west)+(0,0.6)$);
  \node[lbl, font=\footnotesize] at (8.25, 0.1) {context\\embedding};
  \node[ekblock, font=\footnotesize] (ek) at (8.25, -0.6) {$\chunkemb$};
  \node[lbl, font=\footnotesize] at (8.25, -1.3) {chunk\\embedding};
  \draw (8.55, -0.6) -- ($(gen.west)+(0,-0.6)$);

  \draw[-] (11.2,0.8) rectangle (12.3,1.3);
  \node[lbl, font=\footnotesize] at (11.75, 1.05) {$1$};
  \foreach \dy in {-0.13,0,0.13} {\fill (11.75, 0.525+\dy) circle (0.4pt);}
  \fill[black!12] (11.2,-0.25) rectangle (12.3,0.25);
  \draw[-] (11.2,-0.25) rectangle (12.3,0.25);
  \node[lbl] at (11.75, 0.0) {$k$};
  \foreach \dy in {-0.13,0,0.13} {\fill (11.75, -0.525+\dy) circle (0.4pt);}
  \draw[-] (11.2,-1.3) rectangle (12.3,-0.8);
  \node[lbl, font=\footnotesize] at (11.75, -1.05) {$\nchunks$};
  \node[lbl] at (11.75, -1.65) {$\Delta\wma$};

  \draw (gen.east) -- (11.2, 0);

  \node[circle, draw] (plus) at (14.3, 0) {$+$};
  \node[lbl]   (phiout) at (16.2, 0) {$\wma$};

  \draw (12.5,0) -- node[above, font=\footnotesize] {$\Delta\wma$} (plus);
  \node[ekblock, font=\footnotesize] (tmpl) at (14.3, -1.1) {$\wmatmpl$};
  \draw (tmpl.north) -- (plus.south);
  \node[lbl, font=\footnotesize] at (14.3,-1.65) {template};
  \draw (plus) -- (phiout);
\end{tikzpicture}}

%% file: figures/forward-pass.tex
\def\figfont{\fontsize{9pt}{10.5pt}\selectfont}
\begin{tikzpicture}[
  font=\figfont,
  block/.style={draw, rounded corners, fill=white, minimum width=1.6cm, minimum height=0.75cm, align=center, font=\figfont},
  ekblock/.style={draw, rounded corners, fill=trainable, minimum width=0.45cm, minimum height=0.34cm, align=center, font=\figfont},
  db/.style={cylinder, draw, fill=white, shape border rotate=90, aspect=0.22,
             minimum width=2.0cm, minimum height=1.0cm, text width=1.7cm, align=center, font=\figfont},
  lbl/.style={align=center, font=\figfont},
  ->, >=Stealth, very thick
]
  \fill[rounded corners, fill=added] (1.1, -2.15) rectangle (4.5, 4.6);
  \node[lbl, font=\bfseries\scriptsize, anchor=north west] at (2.26, 4.52) {\ourmethod};

  \node[block, minimum width=1.5cm, text width=1.25cm] (mbp)  at (0, 3.5) {Planner $\piupd$};
  \node[db]    (ebuf) at (0, 1.0) {Environment \\ $\bufupd^{j}$};
  \node[block, minimum width=2.0cm, text width=1.5cm] (loss) at (0, -1.3) {Loss \\ function $\Loss$};

  \node[ekblock] (basebox) at (-0.65, 0.0) {$\wmb$};

  \node[block, minimum width=2.4cm, text width=2.0cm] (dcp)  at (2.9, 3.5) {Data collection \\ policy $\picon$};
  \node[db]    (cbuf) at (2.9, 1.0) {Context \\ $\bufcon^{j}$};
  \node[block, minimum width=1.6cm, text width=1.2cm] (hnet) at (2.9, -1.3) {Hypernet \\ $\hnet_{\hparam}$};

  \node[ekblock] (hparambox) at (2.35, 0.0) {$\hparam$};

  \node[lbl] (lossout) at ($(loss.south)+(0,-0.75)$) {$\Loss(\wmb, \wma, \trajupd)$};

  \draw (ebuf) -- node[right] {$\trajupd$} (loss);
  \draw (cbuf) -- node[right] {$\trajcon$} (hnet);

  \coordinate (tickA) at ($(mbp.south)!0.4!(ebuf.north)$);
  \coordinate (tickB) at ($(dcp.south)!0.4!(cbuf.north)$);
  \draw[-] (mbp.south) -- ($(tickA)+(0,0.0675)$);
  \draw ($(tickA)+(0,-0.0675)$) -- (ebuf.north);
  \draw[-] (dcp.south) -- ($(tickB)+(0,0.0675)$);
  \draw ($(tickB)+(0,-0.0675)$) -- (cbuf.north);
  \draw[-, line width=1pt] ($(tickA)+(-0.14,0.0225)$) -- ($(tickA)+(0.14,0.1125)$);
  \draw[-, line width=1pt] ($(tickA)+(-0.14,-0.1125)$) -- ($(tickA)+(0.14,-0.0225)$);
  \draw[-, line width=1pt] ($(tickB)+(-0.14,0.0225)$) -- ($(tickB)+(0.14,0.1125)$);
  \draw[-, line width=1pt] ($(tickB)+(-0.14,-0.1125)$) -- ($(tickB)+(0.14,-0.0225)$);

  \draw[gradflow] (basebox.south) -- (basebox.south |- loss.north);
  \draw[gradflow] (hparambox.south) -- (hparambox.south |- hnet.north);
  \draw[gradflow] (hnet.west) -- node[pos=0.45, above, black, font=\scriptsize] {$\wma$} node[pos=0.45, below, black, align=center, font=\scriptsize] {LoRA \\ adapters} (loss.east);
  \draw[gradflow] (loss.south) -- (lossout);
\end{tikzpicture}

%% file: algorithms/pretraining.tex
\begin{wrapfigure}{r}{0.5\textwidth}
\hrule height.8pt depth0pt \kern2pt
{\captionsetup{font=scriptsize, type=algorithm, justification=raggedright, singlelinecheck=false, skip=0pt}
\captionof{algorithm}{\ourmethod + TD-MPC2 pretraining\label{alg:rollout}}}
\kern1pt\hrule\kern1pt
\begin{algorithmic}[1]
\vspace{-0.2em}
\scriptsize
\Require Environment family $\{E^{1},\,\ldots,\,E^{n}\}$, context-collection policy $\picon$,\;
  initial $\wmb,\,\hparam$,\;
  total training steps $\ntrain$

\Statex
\Statex \textit{// Context data collection}
\hlfortrue
\For{$j = 1, \ldots, n$}
\hlforfalse
  \State \new{Init. and fill $\bufcon^{j}$ by acting in $E^{j}$ with $\picon$}
\EndFor

\Statex
\Statex \textit{// Update data prefill}
\hlfortrue
\For{$j = 1, \ldots, n$}
\hlforfalse
  \State Init. and fill $\bufupd^{j}$ by acting in $E^{j}$ with $\piupd$
\EndFor

\Statex
\Statex \textit{// Online Interaction and Training Loop}
\State \new{$i \;\sim\; \mathrm{Uniform}(\{1,\ldots,n\})$}
  \Comment{Sample rollout env.}
\State $\state \;\leftarrow\; E^{i}.\mathrm{reset}()$
\For{$t = 1, \ldots, \ntrain$}
  \State \new{$k \;\sim\; \mathrm{Uniform}(\{1,\ldots,n\})$}
    \Comment{Sample training env.}
    
    \Statex
    \Statex \hspace{\algorithmicindent}\textit{// Update base and hypernet parameters}
    \State $\trajupd \;\leftarrow\; \mathrm{Sample}(\bufupd^{k})$
    \State \new{$\trajcon \;\leftarrow\; \mathrm{Sample}(\bufcon^{k})$}
  \State \new{$\wma \;\leftarrow\; \hnet_{\hparam}(\trajcon)$}
    \Comment{Training env. adapters}
  \State \new{$\thetaWM \;\leftarrow\; (\wmb,\;\wma)$}
    \Comment{Merge adapters}
  \State $\parens{\wmb,\new{$\hparam$}} \leftarrow \Call{TDMPC2Update}{\trajupd,\new{$\trajcon$},\thetaWM,\new{$\hparam$}}$

  \Statex
  \Statex \hspace{\algorithmicindent}\textit{// Plan and act}
  \State \new{$\trajcon^{i} \;\leftarrow\; \mathrm{Sample}(\bufcon^{i})$}
  \State \new{$\wma \;\leftarrow\; \hnet_{\hparam}(\trajcon^{i})$}
    \Comment{Rollout env. adapters}
  \State \new{$\thetaWM \;\leftarrow\; (\wmb,\,\wma)$}
  \State $a \;\leftarrow\; \piupd{(\state,\;\thetaWM)}$
    \Comment{Plan with latest parameters}
  \State $\state',\; r,\; \mathrm{done} \;\leftarrow\; E^{i}.\mathrm{step}\!\parens{a}$

  \Statex
  \Statex \hspace{\algorithmicindent}\textit{// Update environment and buffers}
  \State $\bufupd^{i} \;\leftarrow\; \bufupd^{i} \cup \{(\state,\;a,\;r,\;\state')\}$
  \If{$\mathrm{done}$}
    \State \new{$i \;\sim\; \mathrm{Uniform}(\{1,\ldots,n\})$}
      \Comment{Sample rollout env.}
    \State $\state \;\leftarrow\; E^{i}.\mathrm{reset}()$
  \Else
    \State $\state \;\leftarrow\; \state'$
  \EndIf
\EndFor

\Ensure $\wmb,\,\hparam$
\end{algorithmic}
\kern2pt\hrule height.8pt depth0pt
\vspace{-25pt}
\end{wrapfigure}

%% file: sections/experiments.tex
\section{Experiments}
\label{sec:experiments}

We test \ourmethod on the environment families shown in \cref{tab:families}, where each family consists of a set of environments that differ in one property.
We use 5 seeds per experiment and report 95\% confidence intervals for the mean across seeds, computed with Student's t distribution. 
\Cref{app:hyperparameters} lists the architecture, optimization, and per-experiment hyperparameters needed to reproduce every experiment below.
We will include a public codebase in the final version of this paper.

\begin{table}[t]
\centering
\scriptsize
\begin{tabular}{@{}lllll@{}}
\toprule
Family & Robot & Objective & Environment property & \# Environments \\
\midrule
Walker & Walker2d       & Walk forward           & Actuator strength   & 20 \\
Reach      & Meta-World arm & Reach target position  & Goal position       & 6 \\
Push       & Meta-World arm & Place puck at goal     & Puck goal position  & 6 \\
Go1 & Unitree Go1 & Run & Running direction (forward or backward)  & 2 \\
Cheetah/Walker & Half-Cheetah / Walker2d & Run forward & Robot embodiment & 2 \\
\bottomrule
\end{tabular}
\caption{Environment families used in our experiments, adapted from DeepMind Control Suite~\citep{tassa2018deepmind} and Meta-World~\citep{yu2019meta}. The environment property column lists the property that varies between environments in a family.}
\label{tab:families}
\vspace{-10pt}
\end{table}

\subsection{Experiment 1: Online Adaptation}
\label{sec:experiment1}

\paragraph{Question.} How does \ourmethod's performance compare to online, gradient-based adaptation?

\paragraph{Setup.}
For each environment in a family, we begin by collecting 3 episodes of context data with $\picon$ (7.5--75s of data depending on the environment).
Then, we roll out a pretrained model, adapting it at every step using the context data.
Gradient-based baselines take one gradient step per environment step, adapting a set of LoRA adapters.
On the other hand, \ourmethod uses its hypernetwork to generate new LoRA adapters at every environment step.
We report episode return (or success rate) averaged across environments and seeds.

\paragraph{Baselines.} We compare \ourmethod against adaptation on models trained on a single task (ST) or with Domain Randomization \citep[DR;][]{tobin2017domain}.
DR pretrains a base model on all environments in the family, and ST pretrains a base model on a single environment in the family.
We also compare against a non-adapting oracle baseline, which appends the environment's ground-truth parameter directly to the state, e.g., actuator strength, or running direction (specified with a one-hot vector).
Since every other method must infer this parameter from context data, the oracle's performance serves as a ceiling on the performance achievable with perfect environment identification.
World model hyperparameters are fixed across baselines.

\begin{figure}[b]
\centering
\includegraphics[width=\textwidth]{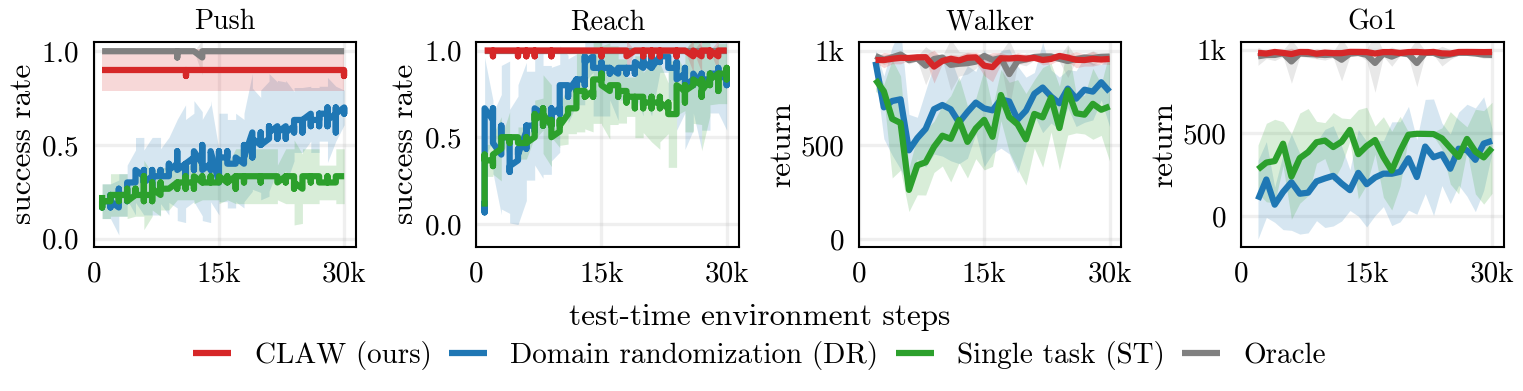}
\vspace{-20pt}
\caption{Experiment 1. Episode return (or success rate) vs. test-time environment steps, averaged over all environments in a family.}

\label{fig:exp1-results}
\end{figure}

\paragraph{Results.} Results are shown in \cref{fig:exp1-results}. 
In all families, \ourmethod closely tracks the oracle's performance.
In the manipulation families (Push and Reach), DR and ST climb toward that ceiling, but they take many environment steps to get there, and performance is very noisy.
In Walker, DR and ST get worse instead of catching up.
In Go1, neither DR nor ST catches up.
DR's low performance is likely due to Go1's two running directions forming a bimodal distribution of behavior, which DR training resolves into a single ``average'' model that is not good at either behavior.
ST pretrains on the forward direction alone, so it starts far from the backward environment and never closes the gap.

\subsection{Experiment 2: Adaptation when Data Is Limited}
\label{sec:experiment2}

\paragraph{Question.} Given that test-time data is often expensive and/or scarce, how does \ourmethod's performance compare to the return achieved by applying many steps of gradient-based adaptation on a fixed test-time dataset?

\paragraph{Setup.} For each family and environment, we collect 3 episodes of context data using $\picon$ (between 7.5 and 75 seconds depending on the environment).
Then, we use this data to adapt all methods offline, freeze the models, and evaluate them.
For the gradient-based adaptation baselines, we sweep a range of offline gradient steps, thereby testing how performance changes as a function of offline adaptation.
We report the same metrics as Experiment 1.

\paragraph{Baselines.} As in Experiment 1, we compare to DR, ST, and an oracle, here also swept over adaptation gradient steps.

\begin{figure}[t]
\vspace{-25pt}
\centering
\includegraphics[width=\textwidth]{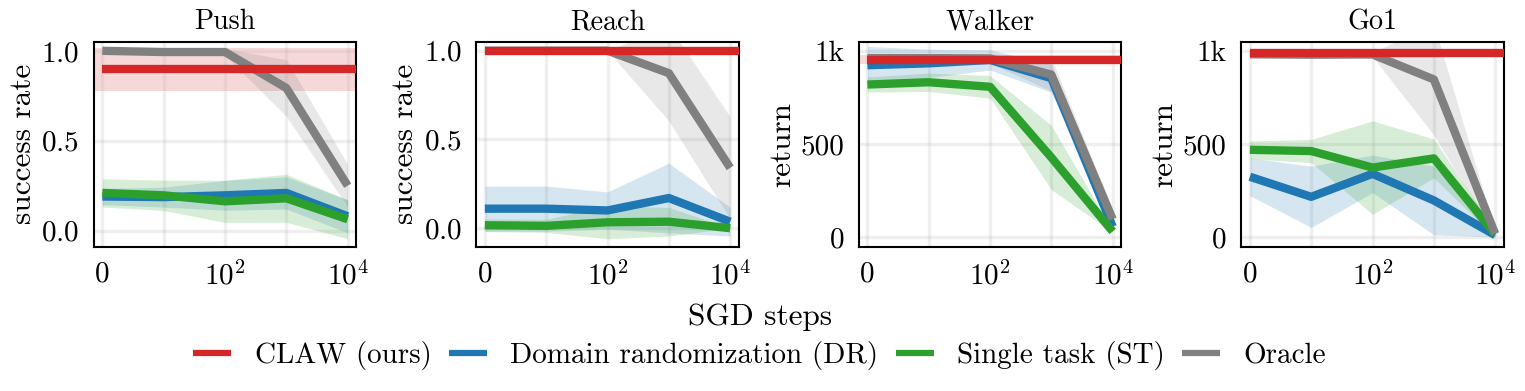}
\vspace{-20pt}
\caption{Experiment 2. \ourmethod's single hypernetwork pass (red) vs. gradient-based baselines (DR, ST, oracle) as a function of SGD steps.}
\label{fig:exp2-results}
\vspace{-5pt}
\end{figure}

\paragraph{Results.} Results are shown in \cref{fig:exp2-results}. 
\ourmethod mostly tracks oracle performance, and outperforms other baselines.
Moreover, as the number of SGD steps increases, performance collapses for all gradient-based methods as they overfit to the data.
This is supported by the collapse of the oracle: since it already knows the environment, the only thing left for it to fit is noise in the small dataset.

\subsection{Experiment 3: Comparison to In-Context Learning}
\label{sec:experiment3}

\begin{wrapfigure}[12]{r}{3.15in}
\centering
\includegraphics[width=\linewidth]{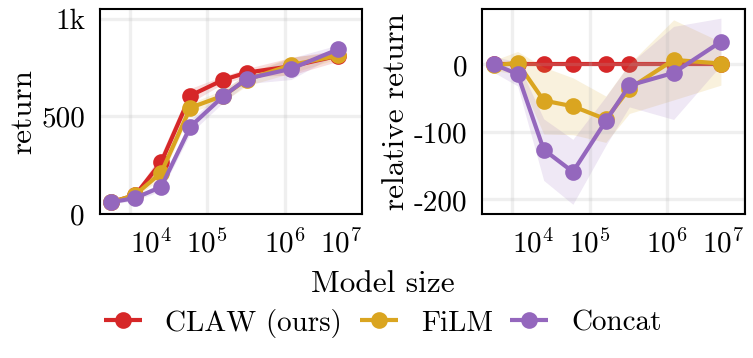}
\vspace{-20pt}
\caption{Experiment 3. \textbf{Left:} return vs. model size. \textbf{Right:} reward relative to \ourmethod.}
\label{fig:exp3-results}
\end{wrapfigure}

\paragraph{Question.} Does \ourmethod's performance depend on generating expressive world-model adapters, or merely conditioning on the context embedding? How does this relationship change with model size?

\paragraph{Setup.} For each environment in the Cheetah/Walker family in \cref{tab:families}, we roll out one episode adapting at every timestep (as in previous experiments).
We choose this family because we need a harder family than the others to tell the in-context baselines apart. 
Cheetah/Walker provides this since state and action spaces differ completely across environments.
We also sweep over model size and report the same evaluation metrics as Experiment 1.

\paragraph{Baselines.} We compare against two in-context learning baselines that condition a fixed model on context, rather than adapting its weights, to isolate the source of \ourmethod's performance.
Concatenation (Concat) feeds the hypernetwork's context embedding directly into the world model's state (instead of using it to generate adapters); this is akin to the approaches of \citet{kumar2021rma,rakelly2019efficient,liu2025locoformer}.
We also compare to a hypernetwork trained to generate FiLM parameters~\citep{perez2017film}, which scale and shift features.

\paragraph{Results.} Results are shown in \cref{fig:exp3-results}.
At the smallest sizes, no method learns enough to separate performance.
At the largest sizes, the base model has enough capacity to represent every environment without conditioning, so performance does not separate either.
\ourmethod's advantage is clearest at intermediate sizes where it outperforms both baselines.
These results align with those of \citet{jayakumar2020multiplicative}, who predict that hypernetwork-generated weights are more expressive than the conditioning mechanisms of FiLM and Concat.

\subsection{Experiment 4: LoRA Rank Sensitivity}
\label{sec:experiment4}

\paragraph{Question.} For a fixed model size, how does \ourmethod's performance change with LoRA rank?

\paragraph{Setup.}
We fix the model size and pretrain a hypernetwork with \ourmethod for a range of LoRA ranks.
We use the Cheetah/Walker family and select an intermediate size where in-context baseline performance is separable.
As LoRA rank increases, we increase the number of chunk embeddings $K$ in \cref{fig:hypernet-forward} and keep all other hyperparameters fixed.
This controls for hypernetwork capacity, so any performance change reflects the rank increase, not a larger generator network.
Then, we evaluate online adaptation performance as in Experiment 1.

\begin{wrapfigure}[17]{r}{1.55in}
\centering
\includegraphics[width=\linewidth]{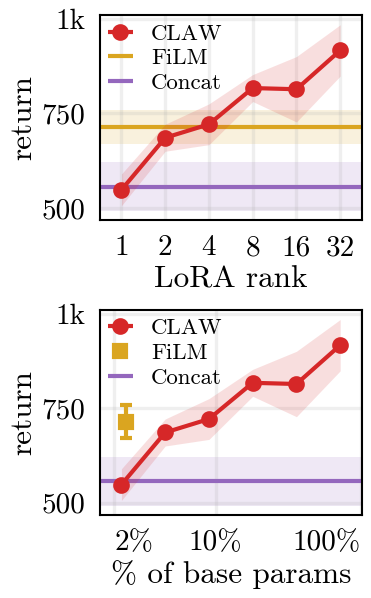}
\vspace{-20pt}
\caption{Experiment 4. \textbf{Top:} return vs.\ LoRA rank. \textbf{Bottom:} return vs.\ adapter size as a percentage of the base parameter count.}
\label{fig:exp4-results}
\end{wrapfigure}

\paragraph{Baselines.} We compare against the same in-context learning baselines as Experiment 3. 

\paragraph{Results.} Results are shown in \cref{fig:exp4-results}.
We report performance vs. adapter rank and adapter size as a percentage of the base parameter count.
\ourmethod's performance rises with rank and percentage.
On the parameter axis, FiLM outperforms \ourmethod at low budgets, and \ourmethod needs around four times FiLM's parameter budget to match its return.
Past that budget, \ourmethod surpasses FiLM, whose performance stays constant since it has no comparable knob to spend a larger budget on.

\subsection{Experiment 5: Post-Hoc Hypernet Comparison}
\label{sec:experiment5}

\paragraph{Question.} How does training the base model and hypernetwork jointly (\ourmethod) compare to training a hypernetwork on top of a frozen pretrained base model \citep[cf.][]{bianchi2026robotic}?

\begin{wrapfigure}[10]{r}{1.55in}
\centering
\vspace{-7pt}
\includegraphics[width=\linewidth]{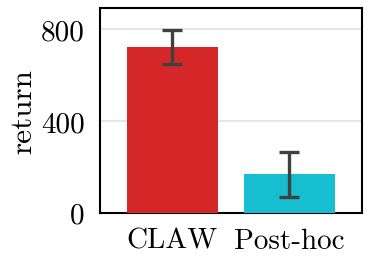}
\vspace{-20pt}
\caption{Experiment 5. Cheetah return for \ourmethod vs. post-hoc baseline.}
\label{fig:exp5-results}
\end{wrapfigure}

\paragraph{Setup.} We evaluate performance on Cheetah using the same metrics as previous experiments.
The post-hoc hypernet's total number of training steps and training data match what \ourmethod would see for this family (500k steps total, split between both environments).

\paragraph{Baseline.} First, we pretrain a plain TD-MPC2 base on Walker only for 250k steps.
Then, we freeze it and train a hypernetwork for 250k steps to generate adapters for the Cheetah environment. We refer to this as the ``post-hoc hypernet.''
This matches the approach proposed by \citet{bianchi2026robotic}, who train a hypernetwork to generate LoRA adapters using a frozen robotics foundation model.

\paragraph{Results.} Results are shown in \cref{fig:exp5-results}. 
\ourmethod outperforms a post-hoc hypernet, showing that jointly training the base model and hypernetwork leads to better performance.

%% file: sections/conclusion.tex
\section{Conclusion, Limitations, and Future Work}
\label{sec:conclusion}

We introduce \ourmethod, a method for world model adaptation that uses a hypernetwork to generate LoRA adapters online, replacing gradient-based test-time adaptation. 
Across all tested environment families, \ourmethod outperforms online gradient-based and in-context learning baselines. 
Moreover, under limited data, \ourmethod avoids the overfitting that degrades gradient-based approaches. 
We also present ablations showing that \ourmethod's performance comes from generating world model adapters rather than merely conditioning on a context embedding, that performance improves with LoRA rank, and that training the hypernetwork jointly with the base model outperforms training it post hoc.

\ourmethod is limited by the identifiability of the environment when using the data-collection policy $\picon$.
Additionally, \ourmethod learns to generate adapters only for the environments seen during pretraining.
Future work includes pretraining and/or architectural improvements for better generalization, applying \ourmethod to sim-to-real transfer, optimizing the data-collection policy for downstream performance, and algorithmically selecting which environments to train on~\citep{dennis2020emergent}.

%% file: sections/acknowledgments.tex
\section*{Acknowledgments}
\label{sec:acknowledgments}

This work was supported by the National Science Foundation under Grant No. 2409535.

%% file: sections/ai_use.tex
\section*{AI Use Statement}
\label{sec:ai-use}

We used generative AI to implement methods and for general coding assistance, to edit and format this paper, and to summarize existing literature and surface relevant work we had not considered.
We did not use generative AI for any other task.
We have reviewed all AI-assisted work.
We take responsibility for the final content of this work, including text, claims or artifacts produced with the aid of generative AI.

%% file: sections/ethics.tex
\section*{Ethics Statement}
\label{sec:ethics}

We are not aware of any ethical concerns raised by this work.

%% file: sections/reproducibility.tex
\section*{Reproducibility Statement}
\label{sec:reproducibility}

\Cref{app:hyperparameters} reports every hyperparameter used to reproduce our experiments.
We will also include a code base with instructions on how to run every experiment in this paper.

%% file: sections/appendix.tex
\clearpage
\appendix
\crefname{appendix}{Appendix}{Appendices}
\crefalias{section}{appendix}

\section{TDMPC2Update: World Model and Policy Update}
\label{app:tdmpc2update}

\Cref{alg:tdmpc2update} is the TD-MPC2 weight update \citep{hansen2024tdmpc2} modified to train the hypernetwork alongside the base model.
It runs in two phases.
The first phase computes the world model loss and updates the world model base slice $\wmb^{\mathrm{wm}}$ jointly with the hypernetwork parameters $\hparam$.
The second phase computes the policy loss and updates the policy base slice $\wmb^{\pi}$, again jointly with $\hparam$.
Between the two phases we regenerate the policy adapter ${\wma^{\pi}}$ with the updated hypernetwork, so the policy loss uses the adapter the current hypernetwork produces.
The update ends by refreshing the target critic and the $Q$-scale.
The target critic EMA tracks base weights only, since its adapter is regenerated on every call.

\begin{algorithm}[h]
\caption{\textsc{TDMPC2Update} --- world model, policy, and hypernetwork update}
\label{alg:tdmpc2update}
\begin{algorithmic}[1]
\Require $\trajupd,\,\trajcon$,\;
  $\thetaWM = (\wmb,\,\wma)$, where $\wma = (\wma^{\mathrm{wm}},\,\wma^{\pi})$ are the world-model and policy adapter slices,\;
  $\hparam$
\State $\z \;\leftarrow\; \mathrm{encode}\!\parens{\trajupd,\,\thetaWM}$
\State $\mathcal{L}_{\mathrm{TDMPC2},\,\mathrm{wm}} \;\leftarrow\; \mathrm{TDMPC2WorldModelLoss}\!\parens{\trajupd,\,\z,\,\thetaWM}$
\State $\wmb^{\mathrm{wm}},\,\wmb^{\pi} \;\leftarrow\; \wmb$
  \Comment{break out world-model and policy slices}
\State $\parens{{\wmb^{\mathrm{wm}}}',\,\new{$\hparam'$}} \;\leftarrow\;
  \Optim\!\parens{(\wmb^{\mathrm{wm}},\,\new{$\hparam$}),\,\nabla_{\!(\wmb^{\mathrm{wm}},\hparam)}\,\mathcal{L}_{\mathrm{TDMPC2},\,\mathrm{wm}}}$
\State $\wmb' \;\leftarrow\; ({\wmb^{\mathrm{wm}}}',\,\wmb^{\pi})$
\State \new{${\wma^{\pi}}' \;\leftarrow\; \hnet_{\hparam'}(\trajcon)$}
  \Comment{Refresh policy adapter}
\State $\thetaWM' \;\leftarrow\; (\wmb',\;\new{$(\wma^{\mathrm{wm}},\,{\wma^{\pi}}')$})$
\State $\mathcal{L}_{\mathrm{TDMPC2},\,\pi} \;\leftarrow\;
  \mathrm{TDMPC2PolicyLoss}\!\parens{\sg{\z},\;Q\!\parens{\sg{\thetaWM'}},\,\thetaWM'}$
\State $\parens{{\wmb^{\pi}}',\,\new{$\hparam''$}} \;\leftarrow\;
  \Optim\!\parens{(\wmb^{\pi},\,\new{$\hparam'$}),\,\nabla_{\!(\wmb^{\pi},\hparam')}\,\mathcal{L}_{\mathrm{TDMPC2},\,\pi}}$
\State $\wmb'' \;\leftarrow\; ({\wmb^{\mathrm{wm}}}',\,{\wmb^{\pi}}')$,\quad
  $\thetaWM'' \;\leftarrow\; (\wmb'',\;\new{$(\wma^{\mathrm{wm}},\,{\wma^{\pi}}')$})$
\State Update target critic: EMA the critic base into a persistent variable; \new{adapter regenerated fresh, not EMA'd}
\State Update $Q$-scale
\Ensure $\wmb'',\,\hparam''$
\end{algorithmic}
\end{algorithm}

\newpage

\section{Hyperparameters and Experimental Details}
\label{app:hyperparameters}

This appendix reports every hyperparameter needed to reproduce our experiments.
\Cref{tab:arch,tab:optim} give the architecture, adapter, optimization, and planning hyperparameters shared by all methods and experiments unless a table below states otherwise.
\Cref{tab:exp12-envs,tab:exp1-protocol,tab:exp2-protocol} cover Experiments 1 and 2, which share pretrained checkpoints across the Walker, Reach, Push, and Go1 families.
\Cref{tab:exp3-sizes,tab:exp4-ranks,tab:exp5-posthoc} cover Experiments 3, 4, and 5, which all use the Cheetah/Walker family.

\subsection{Shared Architecture and Adapter Hyperparameters}
\label{app:arch-hparams}

\begin{table}[H]
\centering
\scriptsize
\begin{tabular}{@{}ll@{}}
\toprule
\textbf{Component} & \textbf{Value} \\
\midrule
\multicolumn{2}{@{}l}{\textit{Encoder}} \\
Hidden features & 256, 512 \\
SimNorm groups / temperature & 8 / 1.0 \\
\addlinespace
\multicolumn{2}{@{}l}{\textit{Dynamics}} \\
Hidden features & 512, 512, 512 \\
SimNorm groups / temperature & 8 / 1.0 \\
\addlinespace
\multicolumn{2}{@{}l}{\textit{Critic}} \\
Hidden features & 512, 512 \\
Ensemble size / subsampled per target & 5 / 2 \\
Return bins / range & 101 / $[-10, 10]$ \\
Dropout & 0.01 \\
\addlinespace
\multicolumn{2}{@{}l}{\textit{Policy}} \\
Hidden features & 512, 512 \\
Log-std range & $[-10, 2]$ \\
\addlinespace
\multicolumn{2}{@{}l}{\textit{Reward}} \\
Hidden features & 512, 512 \\
Bins / range & 101 / $[-10, 10]$ \\
\addlinespace
\multicolumn{2}{@{}l}{\textit{Hypernetwork} $\hnet_{\hparam}$ (\cref{sec:hypernet})} \\
Context dimension $\cdim$ & 64 \\
Transition-embedding MLP $\hemb$ hidden width & 256 \\
Generator MLP $\hgen$ hidden width & 128 \\
Chunk size $c$ (parameters per generated chunk) & 64 \\
Chunk-embedding dimension & 32 \\
\bottomrule
\end{tabular}
\caption{Network architecture, shared by \ourmethod and all baselines (DR, ST, oracle), and by every experiment and family unless overridden in \cref{tab:exp3-sizes,tab:exp4-ranks}. The number of hypernetwork chunks is $\nchunks = \lceil N_{\mathrm{adapt}} / c \rceil$, where $N_{\mathrm{adapt}}$ is the number of parameters in $\wma$; it grows automatically with LoRA rank (\cref{sec:experiment4}).}
\label{tab:arch}
\end{table}

\begin{table}[H]
\centering
\scriptsize
\begin{tabular}{@{}ll@{}}
\toprule
\textbf{Parameter} & \textbf{Value} \\
\midrule
\multicolumn{2}{@{}l}{\textit{TD-MPC2 loss and optimizer} (\cref{alg:tdmpc2update})} \\
Optimizer & Adam \citep{kingma2014adam} \\
Learning rate (world model / encoder / policy) & 3e-4 / 9e-5 / 3e-4 \\
Hypernetwork learning rate (\ourmethod only) & 3e-4 \\
Gradient clip norm & 20 \\
Planning horizon & 3 \\
Discount factor $\discount$ & 0.99 \\
Temporal decay & 0.5 \\
Target critic EMA decay & 0.99 \\
Consistency / reward / value / entropy coefficients & 20.0 / 0.1 / 0.1 / 1e-4 \\
Adapter regularization weight $\regweight$ (\cref{eq:loss-decomp}) & 0.01 \\
Batch size & 256 \\
Initial SGD burst before rollout begins & 1,000 steps \\
\addlinespace
\multicolumn{2}{@{}l}{\textit{Planner (MPPI \citep{williams2015model})}} \\
Horizon & 3 \\
Iterations & 6 \\
Population size & 512 \\
Policy-seeded trajectories & 24 \\
Temperature & 0.5 \\
Action noise std range & $[0.05, 2.0]$ \\
Elites & 64 \\
\bottomrule
\end{tabular}
\caption{Optimization and planning hyperparameters, shared across \ourmethod and all baselines.}
\label{tab:optim}
\end{table}

\subsection{Experiments 1 and 2: Environment Families and Pretraining}
\label{app:exp12-setup}

\begin{table}[H]
\centering
\scriptsize
\begin{tabular}{@{}lrrrrr@{}}
\toprule
Family & $\dim(\state)$ & $\dim(a)$ & Episode length & Envs. (train) & Pretraining steps \\
\midrule
Walker & 24 & 6 & 1000 & 20 & 600,000 \\
Reach  & 39 & 4 & 100  & 6 & 50,000 \\
Push   & 39 & 4 & 100  & 6 & 300,000 \\
Go1 & 39 & 12 & 1000 & 2 & 1,000,000 \\
\bottomrule
\end{tabular}
\caption{Per-family pretraining setup for Experiments 1 and 2, which adapt the same checkpoints. All methods (\ourmethod, DR, ST, oracle) pretrain for the same number of steps per family with 5 seeds each. The oracle baseline appends the ground-truth environment parameter to the state, increasing $\dim(\state)$ by 1 (Walker actuator scale, Go1 running direction) or 3 (Reach/Push goal position).}
\label{tab:exp12-envs}
\end{table}

\begin{table}[H]
\centering
\scriptsize
\begin{tabular}{@{}lll@{}}
\toprule
\textbf{Method} & \textbf{Adapts weights?} & \textbf{Mechanism} \\
\midrule
\ourmethod & no (base frozen after pretraining) & One hypernet forward pass per environment step. Generates rank-16 LoRA adapters.  \\
DR, ST & yes & one gradient step per environment step, on fresh rank-16 adapters \\
Oracle & no & ground-truth environment parameter appended to the state \\
\bottomrule
\end{tabular}
\caption{Experiment 1 online adaptation protocol. Every method rolls out for 30,000 environment steps per environment, with 5 seeds; Optimizer settings for DR/ST's gradient step are in \cref{tab:optim}.}
\label{tab:exp1-protocol}
\end{table}

\begin{table}[H]
\centering
\scriptsize
\begin{tabular}{@{}ll@{}}
\toprule
\textbf{Parameter} & \textbf{Value} \\
\midrule
Context data & 3 episodes, collected with $\picon$ \\
Offline gradient steps swept (DR, ST, oracle) & $\{0, 1, 100, 1000, 10000\}$ \\
\ourmethod adaptation & 1 hypernetwork forward pass (no gradient steps) \\
Evaluation episodes per environment & 10 \\
Seeds & 5 \\
\bottomrule
\end{tabular}
\caption{Experiment 2 offline adaptation protocol. All methods adapt on the same fixed context dataset, then freeze and evaluate. Pretraining is shared with Experiment 1 (\cref{tab:exp12-envs}).}
\label{tab:exp2-protocol}
\end{table}

\subsection{Experiments 3--5: Cheetah/Walker Model-Capacity Studies}
\label{app:exp345-setup}

Experiments 3, 4, and 5 all use the Cheetah/Walker family ($\dim(\state){=}24$, padded for Cheetah; $\dim(a){=}6$; episode length 1000, 2 environments) and share an evaluation protocol: each method re-adapts before every planner call, evaluated over 10 episodes per environment with 5 seeds.

\begin{table}[H]
\centering
\scriptsize
\begin{tabular}{@{}llllrrr@{}}
\toprule
Size & Encoder features & Dynamics features & Critic/Policy/Reward features & Ensemble & LoRA rank & Params.\ $|\wmb|$ \\
\midrule
A & 256, 512 & 512, 512, 512 & 512, 512 & 5 & 16 & 4,960,618 \\
B & 256, 128 & 384, 384, 128 & 384, 384 & 2 & 16 & 1,214,651 \\
D & 192, 48  & 192, 192, 48  & 192, 192 & 2 & 16 & 325,467 \\
E & 128, 32  & 128, 128, 32  & 128, 128 & 2 & 16 & 161,787 \\
F & 64, 32   & 64, 64, 32    & 64, 64   & 2 & 8  & 60,667 \\
G & 32, 32   & 32, 32, 32    & 32, 32   & 2 & 4  & 25,467 \\
H & 16, 32   & 16, 16, 32    & 16, 16   & 2 & 2  & 11,707 \\
I & 8, 32    & 8, 8, 32      & 8, 8     & 2 & 1  & 5,787 \\
\bottomrule
\end{tabular}
\caption{Experiment 3 model-size sweep. Size label C is unused. Params.\ column is the base world model parameter count $|\wmb|$ (excludes hypernetwork parameters $\hparam$), matching the x-axis of \cref{fig:exp3-results}. All other hyperparameters follow \cref{tab:arch,tab:optim}. Sizes F--I additionally shrink LoRA rank, tied across all five adapted modules; FiLM and Concat baselines use the same feature widths with no rank.}
\label{tab:exp3-sizes}
\end{table}

\begin{table}[H]
\centering
\scriptsize
\begin{tabular}{@{}lrrrl@{}}
\toprule
Method & LoRA rank & Adapter params & \% of world model & Return \\
\midrule
\ourmethod & 1  & 1,200  & 2.25\%  & 547.8 $\pm$ 41.9 \\
\ourmethod & 2  & 2,400  & 4.50\%  & 685.8 $\pm$ 35.7 \\
\ourmethod & 4  & 4,800  & 9.00\%  & 721.5 $\pm$ 54.0 \\
\ourmethod & 8  & 9,600  & 17.99\% & 817.3 $\pm$ 35.6 \\
\ourmethod & 16 & 19,200 & 35.98\% & 814.2 $\pm$ 86.6 \\
\ourmethod & 32 & 38,400 & 71.97\% & 917.0 $\pm$ 67.9 \\
FiLM   & --- & 1,280 & 2.40\% & 714.3 $\pm$ 44.3 \\
Concat & --- & --- & --- & 556.2 $\pm$ 64.2 \\
\bottomrule
\end{tabular}
\caption{Experiment 4 rank sweep. Backbone size is fixed at F (\cref{tab:exp3-sizes}); rank is swept jointly across all world model parameters (5 seeds each) while all other hyperparameters stay fixed. Return is mean $\pm$ 95\% CI across 5 seeds, matching \cref{fig:exp4-results}.}
\label{tab:exp4-ranks}
\end{table}

\begin{table}[H]
\centering
\scriptsize
\begin{tabular}{@{}llll@{}}
\toprule
Method & Task(s) & Steps & Trainable \\
\midrule
\ourmethod (joint) & Walker + Cheetah & 500,000 (250,000/task) & base and hypernetwork \\
Post-hoc, pass 1 & Walker only & 250,000 & base only (plain TD-MPC2, no hypernetwork) \\
Post-hoc, pass 2 & Cheetah only & 250,000 & hypernetwork only (base frozen) \\
\bottomrule
\end{tabular}
\caption{Experiment 5 post-hoc comparison, at size F (\cref{tab:exp3-sizes}), rank 8, 5 seeds. \ourmethod's checkpoints are reused from Experiment 3, evaluated on the Cheetah task only to match the post-hoc baseline's single-task evaluation. Both methods see 250,000 steps of Cheetah data and 500,000 total training steps; the only remaining difference is joint vs.\ sequential training of the base model and hypernetwork.}
\label{tab:exp5-posthoc}
\end{table}